# Design Requirements for Human-Centered Graph Neural Network Explanations

**Pantea Habibi**
University of Illinois Chicago
Chicago, IL, USA
phabib4@uic.edu

**Peyman Bagershahi**
University of Illinois Chicago
Chicago, IL, USA
pbaghe2@uic.edu

**Sourav Medya**
University of Illinois Chicago
Chicago, IL, USA
medya@uic.edu

**Debaleena Chattopadhyay**
University of Illinois Chicago
Chicago, IL, USA
debchatt@uic.edu

## Abstract
Graph neural networks (GNNs) are powerful graph-based machine-learning models that are popular in various domains, e.g., social media, transportation, and drug discovery. However, owing to complex data representations, GNNs do not easily allow for *human-intelligible* explanations of their predictions, which can decrease trust in them as well as deter any collaboration opportunities between the AI expert and non-technical, domain expert. Here, we first discuss the two papers that aim to provide GNN explanations to domain experts in an accessible manner and then establish a set of design requirements for human-centered GNN explanations. Finally, we offer two example prototypes to demonstrate some of those proposed requirements.

## Background
Human-computer interaction (HCI) research has become increasingly important in the field of explainable AI (XAI) to meet the various needs of users [8, 23, 11, 4, 5]. These users often belong to different groups—for example, with or without expertise in AI, and with or without domain knowledge—and use different mental models to understand

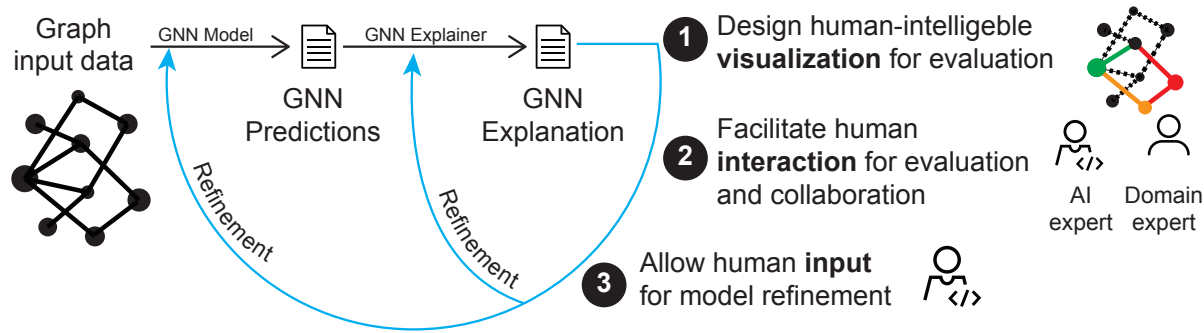

**Figure 1:** We argue that a human-centered approach toward offering Graph Neural Network (GNN) explanations must include human-intelligible visualizations (e.g., a three-dimensional graph with contextual information for selected nodes), a user interface facilitating human interactions (whether AI expert or domain expert) with visualized explanations (e.g., via a virtual reality headset), and finally allow human input during or immediately following such evaluation to refine either the GNN model, explainer, or both.

| Component | DrugExplorer [27] |
|---|---|
| Domain | Drug re-purposing |
| AI area | Knowledge graph |
| Base GNN | Heterogeneous GNN |
| Explainer | GraphMask [20] |
| Objective | Generating tangible explanations |
| Target users | Domain experts in drug repurposing |
| Explanation | GNN explanation |
| Explanation presentation | Providing an overview of all predicted drugs as paths |
| Interactions | Selection, filtering, grouping |
| System design | Helps domain users compare explanation paths at different levels of granularity to generate domain-relevant insights |

**Table 1:** A summary of the primary components of DrugExplorer [27].

XAI [3]. Human-centered XAI (HCXAI) research is particularly lacking for graph AI models [24]. Graph data capture rich structural information among nodes (edges)—and powerful graph-based deep-learning models—called Graph Neural Networks (GNNs)—have garnered significant attention in various domains lately [21, 30, 18]. GNNs can capture relationships within complex networks, such as social networks, and have shown remarkable performance in diverse domains, including drug discovery [29, 21], computer security [18], and recommendation systems [30]. However, like other deep-learning models, GNNs lack transparency—which poses a challenge in their widespread adoption in high-risk domains such as security, healthcare, and finance [31, 10]. To address that, explanations—additional information that describes how important or relevant a feature of the data or input is toward a particular prediction of the model— become useful. Explanations can allow domain experts to understand model characteristics and provide feedback to the AI experts toward improving the model.

While the existing XAI algorithms for GNNs [6] help AI practitioners develop XAI applications, their design choices are mostly driven by technical needs, not end users' explainability needs [8]. For example, GNN explanations are evaluated via quantitative measures such as fidelity [31]. User-based metrics, e.g., perception of the person receiving the explanation, or human-centered evaluations are not used [8, 25]. Thus, a human-centered approach toward designing GNN explanations is needed [8, 11]. While such human-centered efforts to XAI have begun [19, 13, 14, 17, 16, 4], it is clear that there is no one-fits-all solution [8]. For example, unlike textual and visual data, graph data is not easily understandable by humans. Since GNN explanations are often in the form of subgraphs, human-centered evaluation methods for the explanations in textual or image data cannot be directly applied to GNNs [27]. Consequently, a human understanding of GNN explanations remains a challenge. Currently, only two papers discuss techniques to facilitate human understanding of GNN explanations, one focusing on drug re-purposing (DrugExplorer, [27], Table 1) and another on recommendation tasks (RekomGNN, [2],

| Component | RekomGNN [2] |
|---|---|
| Domain | Recommendations |
| AI area | Recommender systems |
| Base GNN | CoreGIE [9] |
| Explainer | None |
| Objective | Understanding predictions |
| Target users | Machine Learning and Software Engineers |
| Explanation | Properties of recommended assets |
| Explanation presentation | Textual info and charts |
| Interactions | Hovering, brushing, rating, drag-and-drop actions |
| System design | Allows users to score the quality of recommendations, and offers a comparative view across different node types and attributes |

**Table 2:** A summary of the primary components of RekomGNN [2].

Table 2). However, they are limited to the type of graphs they tackle, knowledge graphs, and recommender systems, respectively.

DrugExplorer [27] addresses how the visual presentation of GNN explanations impacts user efficiency in accomplishing domain-specific tasks (drug re-purposing). A system with a user interface (UI) offering GNN explanations was designed and evaluated with domain-expert users. Study participants wanted more specific details about the explanations for a deeper understanding and did not find the abstractions useful or aligned with how their typical reasoning process. Although DrugExplorer is among the very few works that take a human-centered approach to GNN explanations, it is limited to the domain it tackles, drug re-purposing, and the type of graph it works with, a knowledge graph.

RekomGNN [2] is another user-facing system that aims to help domain experts evaluate recommendations—in this case, predicted by GNNs. It employs a set of encoding and interaction choices for recommendation tasks, specifically to evaluate the GNN predictions. Predictions are evaluated by their quality or how an input modifies it, while explanations show what data (e.g., part of the recommender graph) is causing that prediction. RekomGNN does not use any GNN explainer, rather, the kind of GNN explanations produced is limited to recommender systems (specific type of graphs with nodes as either users or items to be recommended and are mostly bipartite graphs).

Next, we discuss some design requirements for presenting human-centered GNN explanations.

## HCXAI Design Requirements for GNNs

To make the most out of human involvement with GNN explanations, we need to allow at least three functions (Figure 1):

1. Design *visualization* that is human-intelligible and allows for user evaluation
2. Facilitate *human interaction* with the explanations for evaluation and collaboration
3. Allow *human input* for refinement of GNN model, explainer, or both

While prior work has identified some user-centric requirements for XAI visualization [27], we focus on visualization, interaction, evaluation, and collaboration with GNN explanations. Next, we discuss the *who*, *when*, and *what* of human-centered GNN explanations to establish HCXAI design requirements for GNNs.

### Who: Users

Who would be the users of a GNN explanation, of course, depends on the application domain of the graph data used for training (e.g., molecule), context of the task (e.g., drug discovery), nature of the explanation (e.g., why a drug has certain properties), and expertise. They may either use the AI model or be responsible for improving the model. The existing categorizations of users include people with and without a background in AI, [3], or AI experts, AI novices, and data or domain experts [12]. For GNN explanations, two types of primary users would be *AI experts* and *domain experts*. Domain experts would verify the GNN explanations, provide feedback, and use the GNN predictions. For instance, when GNNs are used in drug discovery [21], the domain experts who have medical expertise would be responsible for validating the explanations. In case of inconsistencies (e.g., an absurd chemical bond as an explanation for a drug), the feedback from medical experts would help AI experts make necessary changes in the GNN model. We also suggest considering two types of secondary users, *organization decision-makers*, who may not use GNN explanations directly but make operational decisions about funding or adoption, and *intermediary operators*, such as machine learning engineers or data scientists.

| What | Details |
|---|---|
| **Operation** | |
| why | reason about why a certain prediction is made |
| why not | reason about why a certain prediction is not made |
| what if | understand how specific modifications will change the prediction |
| how to | investigate the adjustment needed to generate a different prediction |
| what else | query similar instances that generate similar predictions |
| compare in | comparison between explanations of positive and negative examples predicted by the same GNN model |
| compare out | comparison results produced by different GNN models on the same data |
| **Scope** | |
| generic graphs | work for any graph from several domains |
| dynamic graphs | evaluate GNN explanations from dynamic graphs. |

**Table 3:** List of scope and operations in GNN explanations.

*Where and When: Interaction Settings*
Place and time are important design considerations for presenting GNN explanations. We suggest considering two temporal aspects, the *model development timeline*, i.e., when would users engage with GNN explanations, and *collaboration type*, i.e., synchronous or asynchronous. It is also important to consider where user collaboration happens, e.g., *collocated* or *remote*, and how many users are engaging with the GNN evaluations, *one* or *multiple* persons. For example, consider a scenario, where in a drug discovery project, a disease specialist and biochemist (both domain experts but located in different geographic regions) are evaluating GNN explanations along with an AI expert at the same time, before deployment of the model.

*What and How: Operations, Scope, Format, Granularity*
Five types of high-level operations are recommended when designing XAI: why, why not, what if, how to, and what else [7]. In addition, we suggest additional operation requirements for GNN explanations owing to its unique data representation: *compare in* and *compare out* (Table 3). Specific to GNNs, we add two scopes for the operations: (1) generic graphs, and (2) dynamic graphs in any domain. GNNs on dynamic graphs [15] are more complex. A human-centered system should help users quickly evaluate the explanations from different timestamps in the temporal graph and understand how the dynamics evolved toward certain predictions.

The high-level operations mentioned above would need an appropriate output format for the interactions. The usual explanation format for the XAI frameworks is feature importance (attribution), examples (e.g., similar ones), and rules (e.g., decision trees) [26]. However, these formats do not generalize to graph data. In GNNs, the graph structure plays a critical role and most of the GNN explainers [6] produce explanations in terms of nodes, edges, paths, or more broadly, subgraphs. Even if the GNN explainer follows XAI formats such as *rules*, they would still produce the rules combined with subgraphs as concepts [1].

| Interaction | GNN Explanations |
|---|---|
| Select | Node of interest |
| Explore | Positive/negative example |
| Reconfigure | Rotate 3D graphs |
| Encode | Perturb the input graph |
| Abstract/Elaborate | Zoom in and out |
| Filter | Encoded in explanation |
| Connect | Explanations of same positive/negative example |

**Table 4:** A list of low-level interactions, adapted from [28], demonstrating the interaction requirements of GNN explanations.

An HCXAI system needs to facilitate human understanding of such representation of GNN explanations, which is non-trivial for many reasons: (1) *Perception:* Owing to the shape and size, it is not easy to compare two graphs as it is possible to visualize the same subgraphs in different ways (e.g., isomorphic graphs); (2) *Cognitive workload:* Subgraphs usually consist of nodes and edges and the structure is combinatorial, which can be cognitively demanding over time; (3) *Spatial awareness:* Subgraph visualization also requires spatial awareness as several nodes are connected in the subgraph. Thus, 2D visualizations would be limited where 3D visualizations could facilitate additional user interactions (e.g., rotations). Adapting from the information visualization literature [28], Table 4 offers low-level interaction requirements for GNN explanations.

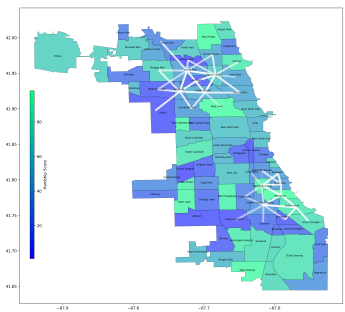

**Figure 2:** A 2D plane graph visualization of GNN explanations on the importance of adjacent neighborhoods for the hardship score prediction of two Chicago neighborhoods.

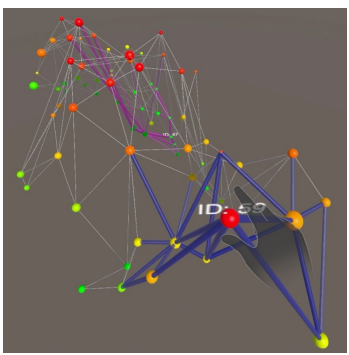

**Figure 3:** An intractable 3D graph depicting 2 selected neighborhoods (ID: 47 "Near North Side" with lowest hardship score of 1 and ID: 59 "Riverdale" with highest hardship score of 98).

**Example Design Prototypes**

We designed two prototypes to demonstrate some of our HCXAI design requirements for GNN explanations (Figures 2 and 3). We used the Chicago neighborhoods dataset [22] to develop a GNN model and predict the hardship score across various Chicago neighborhoods. GNNEXPLAINER [31] was used to obtain the explanations for nodes. The 2D UI (on a 2D map of Chicago) would meet the requirement of visualization, but not interaction or input (Figure 2). Either an AI or a domain expert could understand and evaluate the 2D visualization, but engage with it only in collocated settings (as the UI was designed natively as a desktop GUI). Only the 'why' operations would be possible but the system would work for any generic graphs.

The 3D virtual reality (VR) UI presents the same GNN explanations differently (Figure 3). With this UI, users could select any node in the graph and its explanations. The edge colors would depend on the hardship score of the selected node. The thickness and transparency of the edges would be changed based on the explanation weights. Users also can select multiple nodes and manipulate (rotate, and scale) the graph to compare the selected nodes and their connected edge explanations at the same time. This VR UI would meet the design needs for visualization and interaction but not input for model refinement. Given the affordances of a VR headset, it would allow for synchronous collaboration, in collocated or remote settings (although we did not implement it in our prototype). Compared with the 2D UI, this would allow for 'why', 'why not', 'what if', and 'what else' operations and apply to any generic graphs. Although this UI would allow for interactions, only 'select' and 'explore' interactions would be possible.

The VR UI exemplifies an innovative approach to interacting with GNN explanations by immersing users in a three-dimensional, interactive environment. This not only enhances the intuitiveness and depth of the analytical experience but also leverages the spatial and tactile capabilities of VR technology to facilitate a more comprehensive exploration of complex graph structures and their associated explanations, distinguishing it from common ways of interacting with graph explanations.

In conclusion, we show that despite the interest in human-centered XAI (HCXAI), there is a lack of work in the field of GNN explanations. We contribute to HCXAI by identifying a set of design requirements for human-centered GNN explanations.

Although there are other novel approaches to be explored, such as employing Large Language Models (LLMs) to enhance GNN explanations with text-based narratives for end-users, this work particularly focuses on improving the interpretability and usability of GNN explanations through visual interactions. It aims to understand how tailored interaction modalities and contextualized explanations within Human-Centered Explainable AI (HCXAI) can further bridge the gap between complex GNN outputs and end-user comprehension. This, in turn, has the potential to transform decision-making processes in critical domains. Besides explainability, it would be also interesting to study other critical aspects such as privacy and accountability with similar frameworks. We plan to explore the integration of LLM-based approaches into our current framework as a future direction.

Further, we aim to thoroughly evaluate our user interfaces, which will incorporate the specified design requirements, in real-world scenarios. For instance, we intend to use these user interfaces with GNN models that are deployed on Chicago data for internet equity (Figs. 2 and 3).